\documentclass{article}

\usepackage{PRIMEarxiv}

\usepackage[utf8]{inputenc} 
\usepackage[T1]{fontenc}    
\usepackage{hyperref}       
\usepackage{url}            
\usepackage{booktabs}       
\usepackage{amsfonts}       
\usepackage{nicefrac}       
\usepackage{microtype}      
\usepackage{lipsum}
\usepackage{fancyhdr}       
\usepackage{graphicx}       
\usepackage{multirow}
\usepackage{multicol}
\usepackage{enumitem}
\usepackage{bm}
\setlist{leftmargin=5.5mm}
\usepackage{xcolor}         

\graphicspath{{media/}}     

\title{Mutual Enhancement of Large and Small Language Models with Cross-Silo Knowledge Transfer}

\author{
  Yongheng Deng, Ziqing Qiao, Ju Ren, Yaoxue Zhang \\
  Department of Computer Science and Technology, BNRist \\
  Tsinghua University \\
  Beijing, China\\
  \texttt{\{dyh19,qiaozq20\}@mails.tsinghua.edu.cn} \\
  \texttt{\{renju, zhangyx\}@tsinghua.edu.cn}\\
   \And
  Yang Liu \\
  Institute for AI Industry Research (AIR)\\
  Tsinghua University \\
  Beijing, China \\
  \texttt{liuy03@air.tsinghua.edu.cn} \\
}

\begin{document}
\maketitle

\begin{abstract}
While large language models (LLMs) are empowered with broad knowledge, their task-specific performance is often suboptimal. It necessitates fine-tuning LLMs with task-specific data, but such data may be inaccessible due to privacy concerns. In this paper, we propose a novel approach to enhance LLMs with smaller language models (SLMs) that are trained on clients using their private task-specific data. 
To enable mutual enhancement between LLMs and SLMs, we propose \textsc{CrossLM}, where the SLMs promote the LLM to generate task-specific high-quality data, and both the LLM and SLMs are enhanced with the generated data.
We evaluate \textsc{CrossLM} using publicly accessible language models across a range of benchmark tasks. The results demonstrate that \textsc{CrossLM} significantly enhances the task-specific performance of SLMs on clients and the LLM on the cloud server simultaneously while preserving the LLM's generalization capability.
\end{abstract}

\keywords{Language Models, Collaborative Learning, Cross-Silo, Knowledge Transfer}

\section{Introduction}
\begin{figure}[t]
  \centering
  \includegraphics[width=0.6\linewidth]{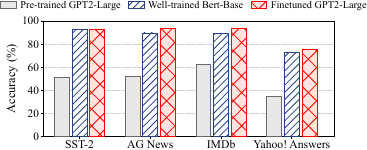}
  \caption{The performance of pre-trained GPT2-Large on specific tasks. The performance of pre-trained GPT2-Large is much degraded relative to a smaller task-specifically trained Bert-Base model. The performance gap can be filled by fine-tuning GPT2-Large with task-specific data.}
  \label{motivation}
\end{figure}

Recent large language models (LLMs) have achieved significant success~\cite{kasneci2023chatgpt,beltagy2019scibert}, sparking a profound revolution in natural language processing (NLP). Today's LLMs are trained on massive corpora from a broad variety of sources, empowering them with diverse and comprehensive linguistic knowledge. However, when pre-trained LLMs are applied to a specific domain for targeted tasks, their performance often proves considerably less satisfactory (see Fig.~\ref{motivation}). Therefore, pre-trained LLMs necessitate further training with domain-specific data to enhance their performance in specialized tasks~\cite{gururangan2020don,hu2021lora}.

Domain-specific data typically originates from specific users, companies, or organizations within a particular field. For example, in the financial sector, banks and investment firms produce transaction records, market reports, and financial statements, which are valuable for tasks such as risk assessment and financial analysis. In the medical domain, hospitals and healthcare providers generate patient medical records and clinical trial data, which are helpful for tasks such as patient diagnosis and medical research. However, this data is usually privacy-sensitive. As user privacy awareness is increasing and legal regulations regarding data governance are becoming more stringent~\cite{regulation2018general}, the collection and utilization of private domain data face significant obstacles. This poses a formidable challenge in further enhancing the performance of LLMs for specific tasks. In this evolving landscape, there is a growing need for innovative approaches that enable the development of domain-specific LLM solutions without exposing private domain data. 

Federated learning (FL)~\cite{mcmahan2017communication}, which harnesses collaborative model training across decentralized data sources without exposing raw private data, provides a viable solution to address these challenges. However, applying FL directly to LLM training encounters obstacles in terms of both resource and proprietary constraints. On the one hand, the training of LLMs is computationally and memory-intensive due to their substantial parameter size, which can be prohibitive for some resource-constrained FL participants. Furthermore, transferring these extensive parameters to the server for aggregation also incurs significant communication overhead. On the other hand, many LLMs are non-public, which cannot be directly accessed or shared with clients for fine-tuning due to proprietary constraints or privacy concerns.

To address the aforementioned challenges, prior arts proposed federated parameter-efficient fine-tuning approaches~\cite{zhao2022reduce, zhang2022federated, babakniya2023slora}. These approaches freeze most of the weights of pre-trained LLMs and only update a small portion of LLM parameters on clients. Only the updated parameters on clients are transferred to the server for aggregation. 
Although this approach reduces the resource overhead compared to full-model training, it still imposes significant resource requirements, as it needs to accommodate the full model on each client, which can be demanding. For example, the GPT-3 model~\cite{brown2020language} boasts a massive 175 billion parameters, necessitating 350GB of GPU memory for parameter storage and inference~\cite{xiao2023offsite}, let alone fine-tuning. Such exorbitant resource requirements may deter a considerable portion of resource-constrained clients from participating in collaborative LLM fine-tuning. To overcome the resource limitations of clients, another line of works proposed to train LLMs in a split way~\cite{tian2022fedbert}, where the computationally intensive portion of LLMs is offloaded to the server for training. However, this approach requires frequent sharing of intermediate representations and/or labels with the server, incurring additional communication costs and potential privacy compromisation of clients. To address the proprietary constraints associated with LLMs, prior efforts such as~\cite{xiao2023offsite} suggested compressing the LLM and appending an adapter to it. Clients then fine-tune the adapter without access to the full model. However, the compressed LLM still has potential proprietary issues.

On the other hand, how to utilize the fine-tuned LLMs for inference is also imperative for clients. The mainstream paradigm is deploying the LLMs on the cloud server, where users send their data to the server for model inference and receive the results afterward. However, this mode places an expensive burden on the server due to the multitude of inference requests, leading to increased costs and response delays, thereby affecting user experience. Moreover, users face privacy risks when uploading their private data to the server. If the model can be deployed on the client-side for local inference, it would effectively address data privacy concerns during inference and alleviate the cost and latency issues associated with centralized inference. Nevertheless, the considerable computational expense of LLMs, even during inference, is not affordable for all clients. Besides, the split inference approach encounters challenges in resolving privacy concerns.

To overcome the challenges associated with LLMs in both training and inference, we propose \textsc{CrossLM}, a novel client-server collaborative training framework for language models. In \textsc{CrossLM}, we assume that the private domain data of clients cannot be shared due to privacy concerns or data regulations. 
To improve the task-specific performance of the server-side LLM, \textsc{CrossLM} proposes to train a smaller language model (SLM) locally on each client using their private task-specific data, and then utilize the SLMs to guide the enhancement of LLM.
Furthermore, the LLM reciprocally enhances the SLMs to compensate for their performance limitations resulting from limited model size or training data.
In this way, \textsc{CrossLM} achieves mutual enhancement of large and small language models with cross-silo knowledge transfer. Notably, considering that clients usually have diverse resource capabilities, \textsc{CrossLM} supports structurally heterogeneous SLMs among clients. As such, clients can customize their SLMs according to their specific resource conditions and needs.

The major technical challenge involved in the design of \textsc{CrossLM} lies in that \emph{how to achieve effective collaborative training between heterogeneous SLMs and the LLM without shared data?} To address this challenge, 
we propose to make full use of the LLM's generative capability to synthesize a task-specific dataset, which is employed to enhance the task-specific performance of the LLM and SLMs. To promote the LLM to generate high-quality data, we propose to make the SLM give feedback on the synthetic data quality and use the feedback as a supervised signal to enhance the LLM. The generated high-quality data is then used to train both the LLM and SLMs. In this way, the LLM and SLMs achieve mutual enhancement in specific tasks.

We evaluate the performance of \textsc{CrossLM} using publicly accessible language models, including GPT2-Large~\cite{radford2019language} and Llama-7B~\cite{touvron2023llama} as our LLMs, and BERT-base~\cite{devlin2018bert} and DistilBERT~\cite{sanh2019distilbert} as our SLMs. We assess their performance on a variety of benchmark tasks from GLUE~\cite{wang2018glue} and SuperGLUE~\cite{wang2019superglue}. The results demonstrate that \textsc{CrossLM} effectively enhances the task-specific performance of SLMs on clients and the LLM on the cloud server simultaneously. Additionally, the results show that \textsc{CrossLM} exhibits only negligible performance degradation on general tasks, verifying its ability to enhance the LLM's task-specific performance while preserving its generalization capacity for a wide range of downstream tasks.

Our contributions are summarized as follows:
\begin{itemize}
    \item To the best of our knowledge, \textsc{CrossLM} represents the first framework that enables collaborative training and mutual enhancement of heterogeneous SLMs on clients and an LLM on the cloud server without sharing clients' private data.
    \item In \textsc{CrossLM}, we propose to leverage SLMs to steer the improvement of the LLM and harness the generative prowess of LLMs to generate synthetic data, facilitating data-free knowledge transfer between the LLM and SLMs. 
    \item We conduct extensive experiments and demonstrate the effectiveness of \textsc{CrossLM} in enhancing the task-specific performance of both SLMs and the LLM while preserving the LLM's generative capabilities. 
\end{itemize}

\section{Related Works}
\label{related_work}
\textbf{Federated Learning Meets Language Models.} Federated Learning (FL) has found extensive applications in the field of NLP. Notably, major technology companies have integrated FL into various products and services~\cite{liu2021federated}. For example, Google has leveraged FL in Gboard mobile keyboard, Pixel phones and Android Messages. Apple employs FL for wake-up word detection in Siri, while doc.ai is developing cross-device FL solutions for biomedical research. Additionally, Snips has introduced cross-device FL for hot-word detection.  
Most previous federated language modeling research applies FedAvg~\cite{mcmahan2017communication} as the federated optimization algorithm~\cite{hard2018federated,ramaswamy2019federated,stremmel2021pretraining,yang2018applied}, where clients train a randomly initialized or pre-trained global model with their local private data, and transfer the model parameters to the cloud server for aggregation. However, it poses unique challenges when applied to large language models (LLMs) due to their colossal parameter sizes, resulting in tremendous communication and computation costs for FL.


\textbf{Federated Fine-Tuning for Large Language Models.} To reduce the communication and computation costs while training LLMs with FL, recent studies have proposed to apply the Parameter Efficient Fine Tuning (PEFT) approach for LLMs~\cite{zhao2022reduce, zhang2022federated, babakniya2023slora}. PEFT offers an alternative training strategy for LLMs, including methods such as Adapter~\cite{houlsby2019parameter}, Prefix-tuning~\cite{li2021prefix}, LoRA~\cite{hu2021lora}, and BitFit~\cite{zaken2022bitfit}. These approaches require fine-tuning only a small subset or additional parameters of LLMs while freezing most or entire pre-trained weights. In~\cite{zhang2022federated}, the authors investigated the application of PEFT methods in FL settings, and demonstrated that PEFT methods can reduce substantial communication overhead in FL settings while still achieving acceptable performance. However, due to the huge parameter size of LLMs, the resource demands of PEFT can be prohibitive for some clients, e.g., edge devices. To address this issue, some works have adopted the concept of split learning. For example, FedBERT~\cite{tian2022fedbert} splits the BERT model into parts and offloads the computation-intensive Transformer layer to the cloud server for training. In this way, each client only requires to train the head layer and embedding layer locally, which can save a huge amount of computation costs. However, in the split learning paradigm, the part of the model offloaded to the cloud server requires updating with the gradient of forward and back propagation from clients. Nonetheless, such a methodology requires sharing intermediate representations and/or labels with the server, which can inadvertently reveal client's input information and potentially undermine the privacy preservation inherent to FL~\cite{niu2022federated}.


\begin{figure}[t]
  \centering
  \includegraphics[width=0.7\linewidth]{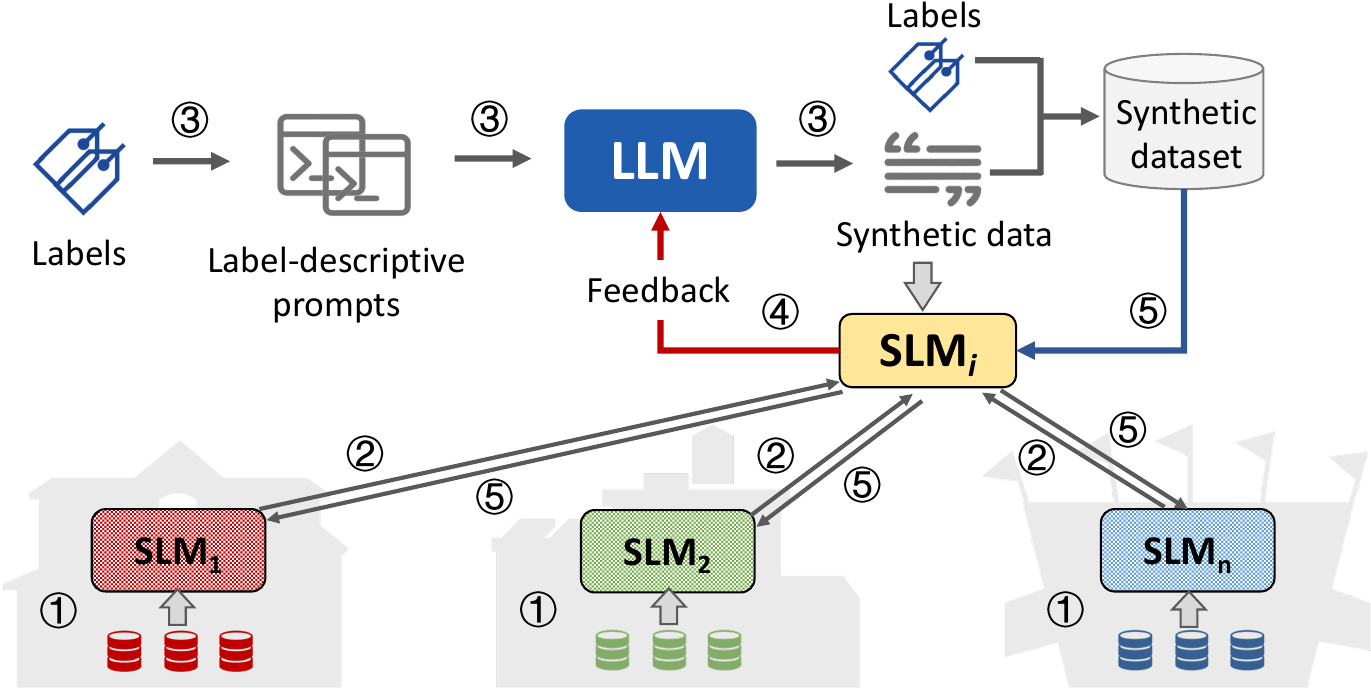}
  \caption{Overview of the \textsc{CrossLM} framework.
  }
  \label{system_overview}
\end{figure}

\section{Problem Setting}
We consider a cross-silo setup with a $C$-class text classification task $\mathcal{T}=\{\mathcal{X}, \mathcal{Y}\}$ with $N$ clients and a cloud server. Each client $i$ has its own private dataset $\mathcal{D}_i$ specific to task $\mathcal{T}$, which is used to train a language model $\mathcal{S}_i$ locally by minimizing the cross-entropy loss function.
On the cloud server, there is a large pre-trained language model $\mathcal{P}$. Since $\mathcal{P}$ is pre-trained with massive corpora, it has the ability of Natural Language Understanding (NLU), which can be used for sentiment analysis in task $\mathcal{T}$. For example, given an input $x_i\in\mathcal{X}$ ``The movie is funny'',  a prompt $\mathcal{M}(\cdot)$ can be instantiated as ``The movie review is'', and a verbalizer $\mathcal{V}(\cdot)$ can map each label $y_i\in\mathcal{Y}$, e.g., ``positive'' or ``negative'', to a word/words in $\mathcal{P}$'s vocabulary. The probability of class $y_i\in\mathcal{Y}$ for $x_i$ is:
\begin{equation}
  p^l(y_i|x_i) = \mathcal{P}(\mathcal{V}(y_i)|\mathcal{M}(x_i)).
\end{equation}
Given $x_i\in\mathcal{X}$, $\mathcal{P}$ outputs a natural language sequence and we aim to maximize $p^l(y_i|x_i)$, i.e., improving the NLU performance of the LLM. On the other hand, we aim to enhance the Natural Language Generation (NLG) performance of $\mathcal{P}$, which is also a fundamental capability of LLMs. That is, given a label-descriptive prompt $\mathcal{M}(y_i)$, e.g., ``Here is a positive movie review:'', we hope $\mathcal{P}$ can generate high-quality sentences in line with the label, for example, ``A good movie!'', rather than ``A bad movie.''.

\section{Methods}
\label{methods}
\subsection{High-level methodology}
We propose \textsc{CrossLM}, a client-server collaborative training framework that trains a large language model (LLM) on the cloud server using the smaller\footnote{"smaller" is relative to the large cloud-side model, a consideration driven by the relatively constrained resources available on clients.} language models (SLMs) from clients trained with their local private data. As shown in Fig.~\ref{system_overview}, the LLM and SLMs are collaboratively trained with the following five steps:

\textbf{Step 1: Each client trains a SLM with their local private data.} To fully utilize the private task-specific data on clients without sharing the raw data, \textsc{CrossLM} proposes to locally train a SLM on clients with their private data. Note that, the SLM on each client can be adaptive to client's resource capabilities with heterogeneous architectures.

\textbf{Step 2: Clients transfer SLMs to the cloud server asynchronously.} Since client's private data is inaccessible to LLM, \textsc{CrossLM} proposes to train the cloud-side LLM with the trained SLMs on clients. Considering that the training speed can be inconsistent between clients, we propose an asynchronous training strategy, i.e., the LLM is trained whenever receiving an SLM from a client.

\textbf{Step 3: Utilizing the generative power of LLM to synthesize a dataset.} To enable knowledge transfer between the LLM and SLM, we fully utilize the generative power of the LLM to synthesize a dataset. \textsc{CrossLM} wraps task-specific labels up into label-descriptive prompts, based on which, the LLM generates synthetic data.

\textbf{Step 4: SLM gives feedback to the synthetic data of LLM.} To promote the LLM to generate high-quality data, each received SLM serves as a task-specific evaluator, which gives feedback on the quality of synthetic data generated by the LLM. The feedback is used as a supervised signal to enhance the LLM.

\textbf{Step 5: SLM is enhanced with the synthetic dataset.} To further improve the performance of SLMs on specific tasks, \textsc{CrossLM} utilizes the generated synthetic dataset to train the SLMs and then returns them to clients for inference.  

\subsection{Knowledge Transfer Between SLMs and LLM}
\textbf{Synthetic Data Generation.} To enable data-free knowledge transfer between the LLM and the SLM, we propose to make full use of the generative power of the LLMs to synthesize a dataset for knowledge transfer. Specifically, the LLM $\mathcal{P}$ will generate a synthetic dataset $\mathcal{D}^s=(\mathcal{X}^s, \mathcal{Y}^s)$ for task $\mathcal{T}$ with label set $\mathcal{Y}=\{y_1, y_2,\dots,y_c\}$. To this end, \textsc{CrossLM} first samples a class label $y^s$ from a uniform distribution:
\begin{equation}
  y^s \sim \bm{U}(y_1, y_2,\dots,y_c).
\end{equation} 
We then wrap up $y^s$ into a label-descriptive prompt $\mathcal{M}(y^s)$, which is input to $\mathcal{P}$ for the generation of $x^s$:
\begin{equation}
  x^s \sim \mathcal{P}(\cdot|\mathcal{M}(y^s)).
\end{equation}
Here, we can adopt some existing strategies such as \cite{fan2018hierarchical,holtzman2019curious} to increase the diversity of the generated dataset.
Afterwards, the generated $x^s$ is paired with $y^s$ to construct the synthetic dataset $\mathcal{D}^s$.

\textbf{Knowledge Transfer from SLMs to LLM.}
To enhance the NLG performance of $\mathcal{P}$, it calls for a feedback on the quality of its generated synthetic dataset. Based on the feedback as a supervised signal, $\mathcal{P}$ can then be enhanced to generate higher-quality natural language data~\cite{ouyang2022training}. Given that each well-trained SLM $\mathcal{S}_i$ exhibits relatively better performance than the LLM $\mathcal{P}$ on a specific task, 
\textsc{CrossLM} proposes to use each $\mathcal{S}_i$ as the feedback provider. Specifically, each generated synthetic data $x^s\in\mathcal{X}^s$ is input to $\mathcal{S}_i$, and then a reward $r(x^s, y^s)$ is calculated based on the output of $\mathcal{S}_i$:
\begin{equation}
  r(x^s, y^s) = 2\mathcal{S}_i(y^s|x^s)-1,
\end{equation}
where $\mathcal{S}_i(y^s|x^s)$ is the probability that the SLM $\mathcal{S}_i$ can classify $x^s$ to $y^s$. Then we design a \emph{quality-driven generation loss} $\mathcal{L}_g$ to promote $\mathcal{P}$ to generate high-quality textual data and suppress it to generate low-quality textual data. $\mathcal{L}_g$ is formulated as:
\begin{equation}\label{loss_generation}
    \mathcal{L}_g = - r(x^s, y^s) \cdot (\log p_{\theta}([\mathcal{M}(y^s), x^s])- \log p_{\theta_{old}}([\mathcal{M}(y^s),x^s])),
\end{equation}
\begin{equation}
    p_{\theta}(s) = \prod_{i}P(s_i|s_{i-k},\dots,s_{i-1};\theta),
\end{equation}
where $\theta$ is the parameters of $\mathcal{P}$ while $\theta_{old}$ is the frozen parameters of the original $\mathcal{P}$. $p_{\theta}(s)$ is likelihood that $\mathcal{P}$ can generate the sequence $s$ given the parameters $\theta$. $r(x^s, y^s)$ is the reward given by a specific SLM $\mathcal{S}_i$. $[\mathcal{M}(y^s),x^s]$ denotes the concatenation of $\mathcal{M}(y^s)$ and $x^s$.
The intuition behind the design of Eq. (\ref{loss_generation}) is to improve the likelihood that the LLM generates high-quality text while diminishing the probability of generating low-quality text. 
Specifically, a large reward ($r(x^s,y^s)>0$) indicates that the generated $x^s$ is a high-quality text. In this case, we aim to increase the probability that $\mathcal{P}$ can generate $x^s$, i.e., making $\log p_{\theta}([\mathcal{M}(y^s), x^s]) > \log p_{\theta_{old}}([\mathcal{M}(y^s),x^s]))$, resulting in a smaller loss value $\mathcal{L}_g$. On the contrary, if the reward is small, diminishing the probability that $\mathcal{P}$ generates the low-quality text $x^s$ results in a smaller $\mathcal{L}_g$. Therefore, the design of Eq. (\ref{loss_generation}) enables smaller loss values $\mathcal{L}_g$ when the LLM has a higher probability of generating high-quality text or a lower probability of generating low-quality text. Consequently, minimizing $\mathcal{L}_g$ encourages the LLM to generate high-quality text and discourages the generation of low-quality text.  

To improve the task-specific NLU performance of $\mathcal{P}$, we propose to utilize the generated synthetic dataset $\mathcal{D}^s$ to train $\mathcal{P}$. Specifically, each synthetic sample $(x^s,y^s)$ is simply transformed by a prompt $\mathcal{M}'(\cdot)$ and then concatenated to a sequence of natural language $[x^s,\mathcal{M}'(y^s)]$. For example, the generated sentence $x^s$=``A good movie!'' for label $y^s$=``positive'' can be concatenated with prompt $\mathcal{M}'(y^s)$=``This is a positive movie review.''. Afterwards, $\mathcal{P}$ is trained with the \emph{standard unsupervised training loss} $\mathcal{L}_t$:
\begin{equation}
    \mathcal{L}_t = - log(p_{\theta}([x^s, \mathcal{M'}(y^s)])),
\end{equation}

\begin{figure}[t]
  \centering
  \includegraphics[width=1\linewidth]{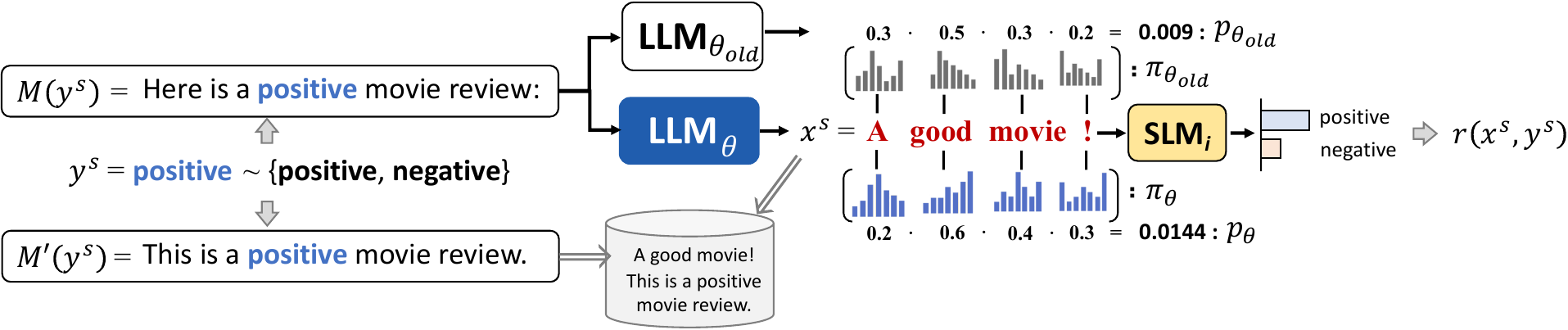}
  \caption{An example illustration of the LLM training process in \textsc{CrossLM}.}
  \label{example}
\end{figure}

While the quality-driven generation loss and the standard unsupervised training loss can enhance the NLU and NLG performance of the LLM on specific tasks, we do not want the LLM to become a specialized model and lose its original generalization capabilities. Therefore, we design a \emph{probability regularization loss} to constrain the parameter changes of $\mathcal{P}$:
\begin{equation}\label{loss_regularization}
    \mathcal{L}_r = D_{KL}(\pi_{\theta}(x^s|\mathcal{M}(y^s)||\pi_{\theta_{old}}(x^s|\mathcal{M}(y^s))),
\end{equation}
where $D_{KL}(\cdot||\cdot)$ is the Kullback-Leibler (KL) divergence, 
$\pi_{\theta}(x^s|\mathcal{M}(y^s)$ and $\pi_{\theta_{old}}(x^s|\mathcal{M}(y^s))$ denote LLM $\mathcal{P}$'s output probability distribution for $x^s$ given $\mathcal{M}(y^s)$ with parameters $\theta$ and $\theta_{old}$, respectively. In Eq. (\ref{loss_regularization}), a lower KL divergence indicates a smaller change in the output of the LLM, serving as a regularization term to limit significant updates on the pre-trained LLM and preserve its original generalization capabilities.

Finally, combining the quality-driven generation loss $\mathcal{L}_g$, the standard unsupervised training loss $\mathcal{L}_t$, and the probability regularization loss $\mathcal{L}_r$, the LLM $\mathcal{P}$ is optimized by minimizing the following objective function:
\begin{equation}
    \mathcal{L}_{LLM} = \gamma_1 \mathcal{L}_g + \gamma_2 \mathcal{L}_t + \gamma_3 \mathcal{L}_r,
\end{equation}
where $\gamma_1$, $\gamma_2$, $\gamma_3$ are hyper-parameters allowing to adjust the relative importance of each component in the overall optimization objective (we set $\gamma_1=1, \gamma_2=0.1$ and $\gamma_3=0.05$ in our evaluations). The design of $\mathcal{L}_{LLM}$ enables to boost LLM's task-specific NLU and NLG abilities while preserving its generalization capacity, ensuring alignment with task-specific instructions. Fig.~\ref{example} illustrates an example of the LLM training process in \textsc{CrossLM}.

\begin{figure}[t]
  \centering
  \includegraphics[width=0.8\linewidth]{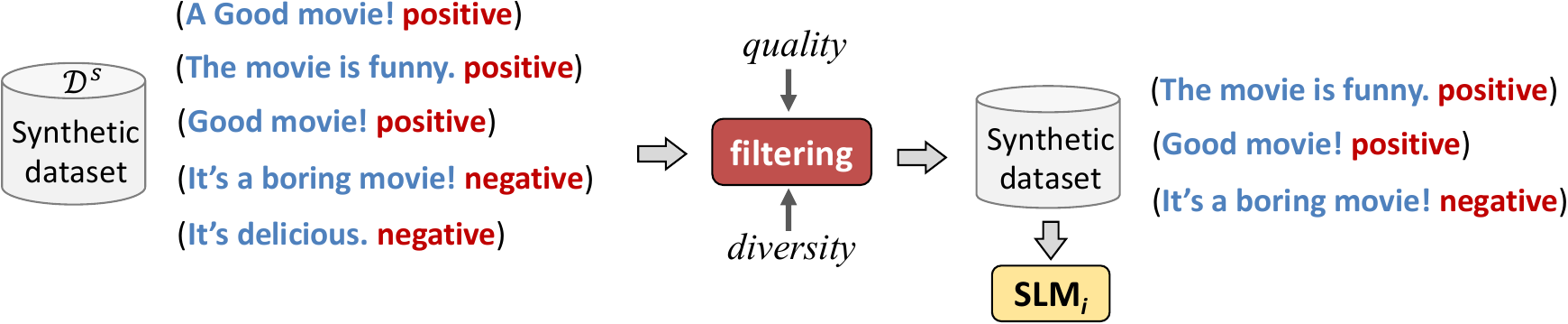}
  \caption{An example illustration of the SLM training process in \textsc{CrossLM}.}
  \label{example_SLM}
\end{figure}

\textbf{Knowledge Transfer from LLM to SLMs.} To further improve the performance of each SLM $\mathcal{S}_i$ on a specific task, we propose to utilize the generated synthetic dataset $\mathcal{D}^s$ to train $\mathcal{S}_i$ on the cloud server. Considering that the quality of the data generated by the LLM is uneven, we propose to first perform a filtering process on the data in $\mathcal{D}^s$ before training $\mathcal{S}_i$. Specifically, we calculate a score for each sample $s_j=(x_j^s, y_j^s)\in \mathcal{D}^s$ based on the \emph{quality} and \emph{diversity} of $s_j$. Likewise, the quality of $s_j$ is evaluated based on the reward $r(x_j^s, y_j^s)$ provided by the SLM $\mathcal{S}_i$. The diversity of $s_j$ is judged by calculating the similarity of $s_j$ with other samples in $\mathcal{D}^s$. To characterize the similarity between $s_j$ and $s_k$, we first transform $x_j^s$ and $x_k^s$ into sentence vectors $v_j=\mathcal{V}(x_j^s)$ and $v_k=\mathcal{V}(x_k^s)$ using an off-the-shelf sentence transformer $\mathcal{V}(\cdot)$~\cite{reimers2019sentence}. Then we calculate their similarity $sim(v_j, v_k)=\frac{v_j\cdot v_k}{||v_j||||v_k||}$, and the diversity $D_j$ of $s_j$ is defined as:
\begin{equation}
    D_j = 1 - \max(\{sim(v_j, v_k)|\forall s_k\in\mathcal{D}^s, k\neq j\}).
\end{equation}
Finally, the score of $s_j$ is defined as the product of $r(x_j^s, y_j^s)$ and $D_j$. Based on the calculated scores, $\alpha\%$ samples in $\mathcal{D}^s$ with the highest scores are selected to train $\mathcal{S}_i$ by minimizing the cross-entropy loss (we set $\alpha\%=25\%$ in our evaluations). Fig.~\ref{example_SLM} illustrates an example of the SLM training process in \textsc{CrossLM}.

\begin{table}[b]
\centering
  \renewcommand{\arraystretch}{1.2}\caption{Information of language models.}\label{model_info}
  \setlength{\tabcolsep}{2.2mm}{
      \begin{tabular}{c|cc|cc}\hline
        & \multicolumn{2}{c|}{\textbf{LLM}} & \multicolumn{2}{c}{\textbf{SLM}}\\\hline
        Model & GPT2-Large & Llama-7B & BERT-Base & DistilBERT \\\hline
        Parameters & 760M & 7B & 110M & 67M
        \\\hline
      \end{tabular}
  }
\end{table}

\section{Experiments}
\label{experiments}
\subsection{Setup}
We set up a scenario with 4 clients and one cloud server, where each client trains an SLM, and a pre-trained LLM is trained on the cloud server.

\textbf{Models} In the evaluations, we set a GPT2-Large~\cite{radford2019language} and Llama-7B~\cite{touvron2023llama} for the LLM, and a BERT-Base~\cite{kenton2019bert} and DistilBERT~\cite{sanh2019distilbert} for the SLM. The information of language models is shown in Table.~\ref{model_info}. Specifically, \textsc{CrossLM} supports heterogeneous model structures between SLMs. In our experimental settings, the SLM on each client is randomly determined as BERT-Base or DistilBERT.

\textbf{Datasets} We evaluate the task-specific performance of the LLM and SLM on four natural language text classification datasets, including SST-2~\cite{socher2013recursive}, IMDb~\cite{maas2011learning}, AG News~\cite{zhang2015character}, Yahoo! Answers~\cite{zhang2015character}. The descriptions and statistics of these datasets are shown in Table.~\ref{dataset_info}. For each dataset, we split its training data following a Dirichlet distribution~\cite{hsu2019measuring} and distribute them to clients as their private domain data.

\textbf{Metrics} Our evaluation spans both the LLM and the SLMs trained within the \textsc{CrossLM} framework. For SLMs, \textsc{CrossLM} aims to enhance their task-specific performance. Therefore, we measure the classification accuracy of each SLM on a specific task, and report the mean accuracy of SLMs. For the LLM, \textsc{CrossLM} aims to improve its NLU and NLG performance on a specific task while keeping its generalization capability. To evaluate the NLU performance of the LLM on a specific task, we report the classification accuracy of the LLM on each specific task. To evaluate the NLG performance of the LLM on a specific task, we make the SLMs to evaluate the quality of the LLM's generated text, i.e., each SLM performs classification on the generated text, and we report the mean classification accuracy of SLMs. To evaluate the generalization capability of the LLM, we select four additional datasets from the GLUE~\cite{wang2018glue} or SuperGLUE~\cite{wang2019superglue} benchmark as illustrated in Table~\ref{NLU_dataset_info}, and measure the performance of the LLM on them after learning on a specific task listed in Table~\ref{dataset_info}. Besides, following previous works~\cite{sun2023simple,frantar2022gptq}, we use the perplexity metric to evaluate the language modeling performance of the LLM, and a lower perplexity indicates better performance.

\begin{table}\centering
  \renewcommand{\arraystretch}{1.2}\caption{Dataset descriptions and statistics for task-specific performance evaluation.}\label{dataset_info}
  \setlength{\tabcolsep}{2.2mm}{
      \begin{tabular}{cccccc}\hline
        \textbf{Dataset} & \textbf{Classification\ Task} & \textbf{\#Class} & \textbf{\#Train} & \textbf{\#Test} & \textbf{Avg\ Length}\\\hline
        SST-2 & Review Sentiment & 2 & 67,349 & 1,820 & 10 \\
        IMDb & Movie Review Sentiment & 2 & 3,600,000 & 400,000 & 78 \\
        AG News & News Topic & 4 & 120,000 & 7,600 & 38 \\
        Yahoo! Answers & Question Answering Topic & 10 & 1,400,000 & 60,000 & 70
        \\\hline
      \end{tabular}
  }
\end{table}

\begin{table}\centering
  \renewcommand{\arraystretch}{1.2}\caption{Dataset descriptions and statistics for LLM's generalization capability evaluation.}\label{NLU_dataset_info}
  \setlength{\tabcolsep}{2.2mm}{
      \begin{tabular}{cccc}\hline
        \textbf{Dataset} & \textbf{Task} & \textbf{\#Train} & \textbf{\#Test} \\\hline
        CoLA & Linguistic Accessibility & 9,594 & 1,063 \\
        MRPC & Paraphrase & 3,668 & 408 \\
        BoolQ & Question Answering & 9,427 & 3,270 \\
        RTE & Textual Entailment & 2,490 & 277
        \\\hline
      \end{tabular}
  }
\end{table}

\subsection{Task-Specific Performance of SLMs}
We first evaluate the task-specific classification performance of SLMs. The performance of \textsc{CrossLM} is compared with two baselines: 1) \emph{Standalone}: each SLM is trained with the corresponding client's local training data without any enhancement from other SLMs or LLM; 2) \emph{Data-free KD}: each SLM is first trained with client's local training data and then enhanced with the synthetic data generated by a pre-trained LLM. Note that the Data-free KD method differs from \textsc{CrossLM} in the utilization of the LLM. Data-free KD employs origin pre-trained LLM for generating synthetic data to enhance the SLMs, while CrossLM utilizes an LLM that has undergone further enhancement. The mean classification accuracy of SLMs trained with different approaches is shown in Table~\ref{SLMs_performance}. It can be observed that the task-specific classification performance of SLMs is improved by an average of $5.8\%-7.8\%$ in Data-free KD and \textsc{CrossLM} compared to the standalone training method. This demonstrates the effectiveness of the LLM in enhancing the performance of SLMs. Moreover, compared to the Data-free KD method, \textsc{CrossLM} also achieves an average of $2\%-2.7\%$ accuracy improvement. This verifies the superiority of \textsc{CrossLM} in promoting the LLM to generate higher-quality text.

\begin{table}\centering
  \renewcommand{\arraystretch}{1.2}\caption{Performance comparison of SLMs with different training approaches.}\label{SLMs_performance}
  \setlength{\tabcolsep}{2.2mm}{
      \begin{tabular}{cccccc}\hline
        \textbf{Method} & \textbf{SST-2} & \textbf{IMDb} & \textbf{AG News} & \textbf{Yahoo! Answers} & \textbf{Average}\\\hline
        Standalone & 0.706 &0.863 &0.819 &0.671 &0.765 \\\hline
         Data-free KD (GPT2-Large) & 0.834 &0.876 &0.839 &0.687 &0.809   \\
         \textsc{CrossLM} (GPT2-Large) & \textbf{0.871} & \textbf{0.882} & \textbf{0.840} & \textbf{0.708} & \textbf{0.825}\\\hline
         Data-free KD (Llama-7B) & 0.796 &0.876 &0.837 & \textbf{0.693} &0.801  \\
         \textsc{CrossLM} (Llama-7B) & \textbf{0.880} & \textbf{0.881} & \textbf{0.840} &0.691 & \textbf{0.823}\\\hline
      \end{tabular}
  }
\end{table}

\begin{table}\centering
  \renewcommand{\arraystretch}{1.2}\caption{Task-specific natural language understanding performance of LLMs after training with \textsc{CrossLM}.}\label{LLM_NLU_performance}
  \setlength{\tabcolsep}{2.2mm}{
      \begin{tabular}{ccccccc}\hline
        \textbf{Model} & \textbf{Method} & \textbf{SST-2} & \textbf{IMDb} & \textbf{AG News} & \textbf{Yahoo! Answers} & \textbf{Average}\\\hline
        \multirow{2}*{GPT2-Large} & Origin & 0.510 &0.524 &0.621 &0.346 &0.500\\
         & \textsc{CrossLM} & \textbf{0.860} & \textbf{0.591} & \textbf{0.826} & \textbf{0.456} & \textbf{0.683}  \\\hline
         \multirow{2}*{Llama-7B} & Origin & 0.610 &0.699 &0.815 &0.530 &0.664\\
         & \textsc{CrossLM} & \textbf{0.908} & \textbf{0.795} & \textbf{0.917} & \textbf{0.580} & \textbf{0.800} \\\hline
      \end{tabular}
  }
\end{table}

\begin{table}\centering
  \renewcommand{\arraystretch}{1.2}\caption{Task-specific natural language generation performance of LLMs after training with \textsc{CrossLM}.}\label{LLM_NLG_performance}
  \setlength{\tabcolsep}{2.2mm}{
      \begin{tabular}{ccccccc}\hline
        \textbf{Model} & \textbf{Method} & \textbf{SST-2} & \textbf{IMDb} & \textbf{AG News} & \textbf{Yahoo! Answers} & \textbf{Average}\\\hline
        \multirow{2}*{GPT2-Large} & Origin & 0.629 &0.663 &0.801 &0.769 &0.716\\
         & \textsc{CrossLM} & \textbf{0.923} & \textbf{0.966} & \textbf{0.976} & \textbf{0.871} & \textbf{0.934}  \\\hline
         \multirow{2}*{Llama-7B} & Origin & 0.658 &0.706 &0.654 &0.787 &0.701\\
         & \textsc{CrossLM} & \textbf{0.808} & \textbf{0.884} & \textbf{0.842} & \textbf{0.826} & \textbf{0.840}  \\\hline
      \end{tabular}
  }
\end{table}

\subsection{Task-Specific Performance of LLM}
Then we evaluate the task-specific performance of the LLM, including the Natural Language Understanding (NLU) performance and the Natural Language Generation (NLG) performance. Table~\ref{LLM_NLU_performance} shows the task-specific classification accuracy of the LLM after training with \textsc{CrossLM}. It can be observed that GPT2-Large and Llama-7B respectively obtain $18.3\%$ and $13.6\%$ absolute accuracy improvement on average compared to its original understanding performance, i.e., prediction accuracy without any task-specific training. The experimental results demonstrate that \textsc{CrossLM} can significantly enhance the NLU performance of the LLM on specific tasks.

To evaluate the NLG performance of the LLM, we show the mean classification accuracy of SLMs on the generated text of the LLM in Table~\ref{LLM_NLG_performance}. It can be observed that the original generation performance of pre-trained LLMs shows significant room for improvement, with only approximately a $70\%$ probability of generating text in line with the specified labels. \textsc{CrossLM} enhances the task-specific NLG performance of GPT2-Large and Llama-7B by an absolute average of $21.8\%$ and $13.9\%$, respectively. The results indicate that \textsc{CrossLM} promotes the LLM to generate higher-quality sentences in line with specific labels.

\subsection{Generalization Capability of LLM}
We next evaluate the generalization capability of the LLM after enhancing its task-specific performance. Table~\ref{generic_LLM_NLU_performance} shows the performance of LLMs before and after being enhanced with \textsc{CrossLM} for specific tasks. We notice that, in most cases, the LLM can still maintain its performance on benchmark datasets after being enhanced for specific tasks, and in the few instances where performance decreases, the decline is minimal. For example, the performance of the pre-trained Llama-7B model improves after being enhanced on the AG News and Yahoo! Answers datasets, and declines slightly on the SST-2 and IMDb datasets. Table~\ref{generic_LLM_NLG_performance} shows the perplexity of LLMs on the Penn Treebank (PTB) dataset~\cite{marcus1994penn} before and after being enhanced with \textsc{CrossLM} for specific tasks. We can see that there is only a small increase in perplexity when the LLMs are enhanced with \textsc{CrossLM} for each specific task. 
These experimental results demonstrate that \textsc{CrossLM} can preserve the generalization capability of LLMs while enhancing their task-specific performance.

\begin{table}\centering
  \renewcommand{\arraystretch}{1.2}\caption{Performance comparison of LLMs on benchmark datasets before and after being enhanced with \textsc{CrossLM} for specific tasks.}\label{generic_LLM_NLU_performance}
  \setlength{\tabcolsep}{2.2mm}{
      \begin{tabular}{ccccccc}\hline
        \textbf{Model} & \textbf{Specific Task} & \textbf{BoolQ} & \textbf{CoLA} & \textbf{MRPC} & \textbf{RTE} & \textbf{Average}\\\hline
        \multirow{5}*{GPT2-Large}  & / & 0.779 &0.822 &0.860 &0.776 &0.809\\\cline{2-7}
        &SST-2 & 0.776 &0.830 &0.863 &0.776 &\textbf{0.811}   \\\cline{2-7}
        & IMDb & 0.785 &0.825 &0.890 &0.787 &\textbf{0.822}  \\\cline{2-7}
        & AG News & 0.783 &0.827 &0.863 &0.765 &\textbf{0.810} \\\cline{2-7}
        & Yahoo! Answers & 0.781 &0.824 &0.878 &0.762 &\textbf{0.811} \\\hline
        \multirow{5}*{Llama-7B} & / & 0.860 &0.864 &0.880 &0.870 &0.868\\\cline{2-7}
        &SST-2 & 0.860 &0.866 &0.873 &0.866 &0.866 \\\cline{2-7}
        & IMDb & 0.862 &0.866 &0.873 &0.863 &0.866 \\\cline{2-7}
        & AG News & 0.860 &0.869 &0.878 &0.874 &\textbf{0.870}  \\\cline{2-7}
        & Yahoo! Answers & 0.864 &0.865 &0.887 &0.870 &\textbf{0.872} \\\hline
      \end{tabular}
  }
\end{table}


\begin{table}\centering
  \renewcommand{\arraystretch}{1.2}\caption{LLMs perplexity results on the PTB dataset before and after being enhanced with \textsc{CrossLM} for specific tasks.}\label{generic_LLM_NLG_performance}
  \setlength{\tabcolsep}{2.2mm}{
      \begin{tabular}{ccccccc}\hline
        \multirow{2}*{\textbf{Model}} & \multirow{2}*{\textbf{Origin}} & \multicolumn{5}{c}{\textbf{Enhanced with \textsc{CrossLM}}} \\\cline{3-7}
        & & \textbf{SST-2} & \textbf{IMDb} & \textbf{AG News} & \textbf{Yahoo! Answers} & \textbf{Average}\\\hline
         GPT2-Large & 19.60 &19.66 &19.69 &19.70 &19.79 &19.71 \\\hline
         Llama-7B & 7.85 & 7.90 &7.88  &7.89 &7.89 & 7.89 \\\hline
      \end{tabular}
  }
\end{table}

\section{Conclusion}
In this paper, we present \textsc{CrossLM}, which can mutually enhance an LLM on the server and SLMs on clients without sharing clients' private data. \textsc{CrossLM} fully utilizes the generative power of the LLM to enable data-free knowledge transfer between the LLM and SLMs. The LLM is enhanced with SLM's feedback while the SLM is enhanced with LLM's generated synthetic data. \textsc{CrossLM} achieves several advantages in a single framework: no need to expose private domain data, reduced resource demand for client computation, heterogeneous model architecture support for clients, asynchronous and one-shot federated learning. Extensive experiments show that \textsc{CrossLM} can improve the task-specific performance of the LLM and SLMs while preserving the generalization capability of the LLM simultaneously.

\bibliographystyle{unsrt}  
\bibliography{CrossLM}

\end{document}